\documentclass{article}

\usepackage{arxiv}

\usepackage[utf8]{inputenc} 
\usepackage[T1]{fontenc}    
\usepackage{hyperref}       
\usepackage{url}            
\usepackage{booktabs}       
\usepackage{amsfonts}       
\usepackage{nicefrac}       
\usepackage{microtype}      
\usepackage{lipsum}
\usepackage{fancyhdr}       
\usepackage{graphicx}       
\graphicspath{{media/}}     
\usepackage{natbib}
\usepackage{doi}
\usepackage{mathtools}

\usepackage{enumitem}

\usepackage{array}       
\usepackage{ragged2e}
\renewcommand{\headeright}{A Preprint}

\title{Agentic Problem Frames: A Systematic Approach to Engineering Reliable Domain Agents}

\author{
  Chanjin Park \\	
  Institute of Engineering Research\\
  Seoul National University\\
  Seoul, Korea \\
  \texttt{calling6@snu.ac.kr} \\
}

\begin{document}
\maketitle

\begin{abstract}

Large Language Models (LLMs) are evolving into autonomous agents, yet current ``frameless'' development—relying on ambiguous natural language without engineering blueprints—leads to critical risks such as scope creep and open-loop failures. To ensure industrial-grade reliability, this study proposes \textbf{Agentic Problem Frames (APF)}, a systematic engineering framework that shifts focus from internal model intelligence to the structured interaction between the agent and its environment.

The APF establishes a \textbf{dynamic specification paradigm} where intent is concretized at runtime through domain knowledge injection. At its core, the \textbf{Act-Verify-Refine (AVR) loop} functions as a closed-loop control system that transforms execution results into verified knowledge assets, driving system behavior toward \textbf{asymptotic convergence} to mission requirements ($R$). To operationalize this, this study introduces the \textbf{Agentic Job Description (AJD)}, a formal specification tool that defines jurisdictional boundaries, operational contexts, and epistemic evaluation criteria.

The efficacy of this framework is validated through two contrasting case studies: a delegated proxy model for business travel and an autonomous supervisor model for industrial equipment management. By applying AJD-based specification and APF modeling to these scenarios, the analysis demonstrates how operational scenarios are systematically controlled within defined boundaries. These cases provide a conceptual proof that agent reliability stems not from a model’s internal reasoning alone, but from the \textbf{rigorous engineering structures} that anchor stochastic AI within deterministic business processes, thereby enabling the development of verifiable and dependable domain agents.

\end{abstract}


\section{Introduction}

\subsection{Background: The Risks of Frameless Development and the Necessity of Engineering}

Artificial Intelligence has evolved beyond simple chat interfaces into the era of \textbf{Agentic AI}, where models are integrated with legacy systems and physical equipment to perform practical tasks. This shift implies that AI has emerged as a specialized workforce within industrial sectors. However, while the pace of technological advancement has been rapid, engineering methodologies have lagged behind. Consequently, current development remains in a \textbf{frameless} state—lacking the structural discipline required to ensure the quality of individual components within a complex system.

From a software engineering perspective, for an agent to function as a reliable engineering component, its boundaries and input/output must be strictly defined, even if its internal logic remains a black box. 
In this study, such structuring begins with the establishment of a Job Description (JD). Developing without a JD is akin to deploying a highly educated but inexperienced new employee to a field without any guidance. 
An agent, like a new recruit with strong academic background (general intelligence) but no domain-specific experience, requires a clear definition of its mission, authority, and work environment. 
Furthermore, the JD must specify the reference materials for the task and the criteria by which the performance will be evaluated. 
When combined with the inherent stochasticity of agents, the absence of this foundational structure leads to the following critical risks:

\begin{itemize}

    \item \textbf{Absence of Boundaries (Scope Creep):} This is akin to a ``recruit who does not know the line''. LLMs possess vast general knowledge. Without clearly defined jurisdictional boundaries, an agent may rely on general common sense to access systems beyond its authority or interfere with unnecessary domains. When accidents occur, a \textbf{responsibility gap} arises; it becomes unclear whether the failure was the model's reasoning error or the manager's failure to provide explicit regulations and work permissions.

    \item \textbf{Absence of Grounding (Knowledge \& Reality Gap):} 
    This resembles a ``recruit who works by guesswork''. Agents often fill in incomplete or ambiguous inputs with probabilistic guesses. Without domain knowledge ($K$) provided as a baseline to anchor the user's intent, the agent will formulate erroneous plans based on hallucinated assumptions. This lack of initial grounding creates a disconnect between the user's actual requirement and the agent's internal reasoning, rendering the starting point of the mission unreliable.
            
    \item \textbf{Absence of Feedback (Open-Loop):} 
    This is the case of a ``recruit who skips the engineering report''. Without a systematic callback mechanism or human confirmation, the agent falls into an \textit{open-loop state}, relying on the flawed assumption that ``execution equals success.'' This leads to a severe \textbf{state mismatch} between the actual world and the agent's internal assumptions, especially when an agent concludes task completion based solely on a successful API call without confirming the actual state change. Furthermore, the lack of feedback prevents the system from assetizing execution results into verified knowledge, thereby hindering the agent's ability to refine its subsequent actions through asymptotic convergence.

\end{itemize}

Recent research by \citet{huang2025professional} underscores the necessity of addressing this engineering crisis. Their study indicates that experienced developers tend to reject "vibe coding"—which delegates full authority to AI—and instead prefer to strictly control AI through explicit context provision (input) and rigorous result verification (output). This paper responds to these industrial requirements and aims to transform frameless development into a controllable engineering process by defining structured boundaries for both input and output

\subsection{Proposed Approach: Structuring via Agentic Problem Frames}
Modern Software Engineering (SE) and Requirements Engineering (RE) have primarily focused on deriving the specification ($S$) for the Machine ($M$) to satisfy requirements ($R$), while ensuring that the implementation aligns with the domain knowledge ($K$) concerning the World ($W$). In this paradigm, the central concerns are: “Has the user's need been accurately translated into a spec?'' and “Is the software implemented without errors according to that spec?''.

However, this implementation-centric approach is no longer valid in an agentic environment because the internal logic of the machine ($M$) is a black box that operates probabilistically. Unlike traditional software, specifications for agents cannot be fixed at design time. This is because trigger events, such as a user’s vague request to "prepare for a business trip" or an ambiguous anomaly signal from a motor, are inherently incomplete. To bridge this gap, the specification must manifest as a Dynamic Specification ($S_t$), concretized just before execution by injecting domain-specific context—such as corporate policies, technical manuals, or historical data.

Consequently, rather than attempting to tune the uncontrollable internals of the agent—such as through model fine-tuning—this study focuses on strictly structuring the external world and interaction mechanisms. By bounding the agent’s reasoning within a professional jurisdictional context through an Agentic Job Description, this study transforms a general-purpose probabilistic model into a reliable engineering component capable of fulfilling specific missions.

To address the inherent uncertainty of agents, this study adopts Michael Jackson’s Problem Frames (PF) theory as its theoretical foundation. PF shifts the focus from the machine’s internal code to the interaction between the machine ($M$) and the world ($W$), expressed through the \textbf{requirement satisfaction argument}: $S, K \vdash R$. This reasoning implies that even if $M$ is an uncontrollable black box,the proposed methodology enables the satisfaction of requirements ($R$) by rigorously designing the specifications ($S$) through which the machine meets the world, while accurately reflecting domain knowledge ($K$).

Building upon this, this paper proposes Agentic Problem Frames (APF)  to transform probabilistic machines into controllable engineering components. The core logic of APF is to achieve requirement satisfaction not through a single static proof, but through \textbf{asymptotic convergence} driven by three mechanisms:

\begin{itemize}

    \item  Dynamic Specification ($S_t$): At runtime, ambiguous user requests are complemented with jurisdictional context to bound the machine's reasoning range.

    \item  Knowledge Transformation ($K_{t+1}$): Opaque state changes in the world are converted into confirmed knowledge through a rigorous verification process.

    \item  Feedback Loop: This newly acquired knowledge is fed back as context for the next cycle, allowing the system to incrementally satisfy $R$ through iterative evolution.

\end{itemize}

For the practical development of such agents, the Agentic Job Description is introduced as the foundational engineering specification. Just as a professional contract defines the mission and authority of a human recruit, the AJD functions as a blueprint that prescribes the jurisdictional boundaries, operational contexts, and epistemic evaluation criteria for the agent. By defining these discrete units through the AJD, this approach transforms a stochastic model into a mission-oriented \textbf{Job Performer}, ensuring that its autonomous judgment is anchored within the APF framework.

This approach represents a paradigm shift from traditional internal code correctness toward fitness, demonstrating how the evolutionary interaction between the machine and the world ultimately ensures reliability in complex industrial domains.

\section{Background and Related Work}
\subsection{Problem Frames Theory and Structuring Agentic Environments}
This study adopts the Problem Frames theory, pioneered by Michael Jackson and Pamela Zave \citet{jackson2001problem}, as its theoretical foundation. Unlike traditional requirements engineering methods that focus on specifying the machine's internal functions or behavioral requirements, PF emphasizes modeling the problem space itself.

The essence of PF lies in shifting the focus from the machine's internal implementation to the interaction between the machine ($M$) and the world ($W$). In an agentic environment, where the machine is an autonomous black box performing probabilistic execution, this interaction-centric view is essential. By leveraging PF, the objective is not to specify the uncontrollable internal logic of the Large Language Model (LLM); instead, this framework \textit{tames its stochastic behavior} by strictly structuring the shared phenomena through which the agent meets the world. This approach enables the transformation of a general-purpose probabilistic model into a reliable engineering component by grounding its reasoning within domain-specific knowledge ($K$)

\subsubsection{The Essence of Problem Frames: Structuring the World, Not the Machine}

The core of Jackson's PF theory lies not in designing the internals of the machine ($M$), but in structuring the world ($W$) with which the machine interacts. \citet{zave1997four} criticized the common tendency to focus exclusively on code—neglecting the environment—as \textbf{implementation bias}. They defined requirements ($R$) not as the machine’s internal behavior, but as the "desirable state changes to be achieved in the world ($W$) through the machine’s actions". For instance, sending a control signal is merely a machine specification ($S$), while the result—"the elevator safely arriving at the desired floor"—is the actual requirement ($R$) fulfilled in the world.
This relationship is demonstrated through the requirement satisfaction argument: $S, K \vdash R$

This formula implies that the specification ($S$) alone cannot satisfy the requirement ($R$); it must be combined with domain knowledge ($K$). While this theory remained a minority view in mainstream software engineering for the past 30 years, it was largely because traditional digital environments were assumed to be causal and deterministic. In such a world, developers could achieve system stability through internal correctness—operating under the optimistic assumption that once a command is correctly issued by the machine, the world will invariably react as designed.
However, the rise of Agentic AI reverses this situation. In this new paradigm, the machine (an LLM) is a black box that perceives and communicates through ambiguous natural language, rather than deterministic logic. Unlike traditional systems, a probabilistic agent cannot internally guarantee the causal outcomes of its actions in a complex world. Without a feedback mechanism to verify if execution results align with $R$, an agent faces structural limitations in correcting its probabilistic judgments.
Thus, controlling this uncertainty necessitates the injection of domain knowledge ($K$) as external context and the physical verification of state changes. These verified changes are then assetized as new context, forming a dynamic cycle that enhances the consistency of subsequent actions. Ultimately, structuring the world and implementing post-verification mechanisms are essential to preventing hallucination.

\subsubsection{Dependability via Decomposition: Isolating Uncertainty}

To solve complex software problems, \citet{jackson2004problem} emphasized decomposing the machine into verified "known problem classes" rather than treating it as a big monolithic entity. He identified \textbf{dependability exposure}—a state where a machine operates based on false assumptions about the world without verification—as the fundamental cause of system failure.
This insight accurately foretells the engineering risks of modern LLM agents. A big monolithic agent, lacking structured boundaries, is likely to mobilize vast, irrelevant knowledge without context, leading to erroneous execution.

If the scope of the world an agent must handle is too broad, the agent inevitably relies on numerous uncertain assumptions. The engineering essence of decomposition is  \textbf{to isolate the world's uncertainty} and minimize the number of assumptions, thereby ensuring a state where each action can be objectively verified. 
Following the principle of isolation through decomposition, this study employs the Agentic Job Description to break down ambiguous agent roles into clear job units. By defining these discrete units, the propagation of uncertain assumptions is physically blocked ensuring that each agent functions as a reliable, verifiable component. 

This modular approach allows for the composition of complex, stable organizations from atomic unit agents. When individual agents are reliably defined as mission-oriented components, the overall system can scale into a coordinated "workforce" or agentic organization, where the reliability of the whole is grounded in the verified performance of its constituent units.

\subsubsection{Classification of the World: Domain Natures and Control Strategies}

Structuring the world begins with identifying the \textbf{nature of the domains} with which the machine interacts. Control strategies and verification methods must fundamentally differ depending on whether a target follows physical/logic laws, possesses autonomy, or consists of structured information. \citet{jackson2001problem} categorized these based on their \textbf{response characteristics}:

\begin{itemize}
    \item \textbf{Causal Domains (Deterministic Interaction)}: These environments show predictable reactions according to physical laws or designed logic (e.g., hardware sensors, internal APIs). While they cannot "refuse" a command, the agent must not assume success based solely on a sent signal. For instance, after a "A public transportation booking  API call, the agent must perform \textbf{technical verification}—such as monitoring confirmation emails or checking status callbacks—to ensure the intended state change actually occurred in the world.
    \item \textbf{Biddable Domains (Autonomous Interaction)}: These are non-deterministic environments with their own autonomy or business constraints, most notably human users. A biddable domain can refuse requests based on internal states or preferences. Since the machine cannot force execution, the strategy must be \textbf{bidding}—proposing an action and waiting for \textbf{human confirmation}. For example, the final itinerary comprising transportation and lodging should only be finalized after a human's "value confirmation."
    \item \textbf{Lexical Domains (Static Information)}: These consist of static data such as manuals, jurisdictional context, or historical logs. In an agentic environment, these are not targets of control but \textbf{sources of knowledge ($K$)}. The control strategy here focuses on precise retrieval; these domains must be injected as context to ground the agent's reasoning, ensuring that the "probabilistic guess" is replaced by "domain-specific judgment."
\end{itemize}

\subsection{Trends in LLM Agent Engineering Research}

Recent research on agents has focused on enhancing internal reasoning capabilities by reflecting the results of interactions with the world. However, engineering approaches to control these agents as reliable, predictable systems have been not enough. Agents in industrial settings must not be mere consumers of general knowledge; they must become engineering entities that reliably execute specific missions within strictly defined boundaries.
Therefore, modeling and structuring the world—specifically the rules and boundaries required for an agent’s mission—must precede unstructured adaptation or trial-and-error learning. Without a predefined problem frame, an agent's reasoning remains ungrounded, leading to unpredictability in complex workflows. This section reviews major prior works and discusses how the proposed APF complements their limitations to ensure agent reliability through structural rigor.

\subsubsection{Challenges in AI Systems Engineering: Ambiguity in Specifying Desired Outcomes}

Attempts to handle AI from a traditional Software Engineering (SE) perspective have persisted, yet a gap between theory and practice remains. \citet{bosch2021engineering} identified through industrial case studies that the primary bottleneck in moving ML models to production lies in the ambiguity of \textbf{``specifying desired outcomes.’'}

The essence of this ambiguity lies in the fact that agent behavior is governed by probabilistic inference rather than deterministic logic. Unlike traditional software, where a specification ($S$) serves as a static blueprint for code, a specification for an agent cannot be fully fixed at design time. This limitation arises because trigger events—such as vague user requests or unpredictable environmental signals—are inherently incomplete. To bridge this gap, the specification must evolve into a \textbf{dynamic specification ($S_t$)}, which is manifested at the moment of execution ($t$) by augmenting ambiguous inputs with the domain knowledge available at that time ($K_t$).

This research addresses these challenges by proposing a structure that clearly delineates an agent’s jurisdictional boundaries via Agentic Job Description. By assetizing execution results into verified domain knowledge for continuous feedback, this framework transforms initially uncertain executions into a process that incrementally satisfies requirements ($R$) through iteration. This approach reframes the specification challenge from a problem of "initial correctness" to one of \textbf{dynamic grounding and verification}.

\subsubsection{Combining Reasoning and Acting: Toward Verified Grounding}

In the evolution of agent architectures, Chain-of-Thought (CoT) research has provided a foundational milestone by proving that LLMs can perform deep reasoning through step-by-step cognitive processes \citep{wei2022chain}. This internal reasoning capability is a crucial component of modern agents. Building upon this, the ReAct framework demonstrated a powerful paradigm of interleaving reasoning and acting, allowing agents to dynamically adjust their plans based on observations from the world \citep{yao2023react}.
While acknowledging the significant contributions of these approaches, this study identifies an engineering gap in the way observations are handled. In the ReAct paradigm, environmental feedback is primarily processed as natural language context. From a systems engineering perspective, this relies on the model’s internal consistency to interpret results correctly, which may be vulnerable to hallucinations when dealing with complex domain logic.
The contribution of APF lies in complementing these reasoning-centric frameworks with an engineering-centric \textit{closed feedback loop}. Rather than treating environmental feedback as mere text to be read, APF introduces a \texttt{Verify} function that transforms state changes into \textit{verified knowledge ($\Delta K$)}. This ensures that the agent's deep reasoning (CoT) is not performed in a vacuum but is grounded in a verified reality. By assetizing results into existing domain knowledge ($K$), APF provides the structural rigor necessary to \textbf{tame} probabilistic execution into a reliable industrial process.

\subsubsection{Tool Use and Reflection: Establishing Reference Points for Incremental Convergence}

Research such as \textbf{Reflexion} by \citet{shinn2023reflexion} proposed a self-reflection mechanism where agents learn from past failures through linguistic feedback. Similarly, approaches like \textbf{Toolformer} \citet{schick2023toolformer} enable language models to autonomously teach themselves to use external tools through self-supervised experiences. These advancements suggest that agents can be optimized through experience rather than fixed code. However, effective reflection and autonomous tool use require a clear reference point—a grounded model of the world—to define what constitutes a "correct" or "desirable" outcome. Without this, self-reflection risks becoming a subjective loop of misinterpretation, and unguided tool use may lead to unverified or uncontrolled side effects.

This study concretizes the necessity of such a reference point through the AVR (Act-Verify-Refine) loop. Specifically, a closed-loop structure is proposed where the agent’s actions and tool invocations are evaluated against the mission goals and domain constraints defined in the AJD. This approach transforms the agent's adaptation from ``vague improvement'' into a rigorous process of asymptotic convergence toward requirement satisfaction ($R$). In this framework, each iteration of the AVR loop reduces the delta between the current state and the requirements, ensuring that the agent's tool use and subsequent refinements are not random but are steered by the structural rigor of the Agentic Problem Frames.

\subsubsection{Multi-Agent Systems: Composing Organizations from Reliable Components}

Beyond single-agent systems, frameworks such as \textbf{Generative Agents} \citep{park2023generative} and \textbf{AutoGen} \citep{wu2023autogen} have explored scenarios where multiple agents collaborate to solve complex problems through natural language dialogue. While recent approaches like \textbf{ChatDev} \citep{qian2024chatdev} and \textbf{MetaGPT} \citep{hong2024metagpt} have attempted to structure these interactions using waterfall models or Standardized Operating Procedures (SOPs), they still largely rely on "chat-based" consensus, which often leads to cascading hallucinations and uncontrollable emergent complexity \citep{he2025llm}.

This study argues that a stable agentic organization cannot be formed by simply connecting conversational agents; it must be composed of \textit{rigorously defined engineering components}. Drawing on the \textit{Single-Responsibility Principle} for production-grade agents \citep{bandara2025practical}, APF focuses on the internal and external structuring of the \textbf{unit agent} via the \textbf{Agentic Job Description} (AJD). By transforming a stochastic model into a mission-oriented component with distinct jurisdictional boundaries and verifiable interfaces, APF provides the fundamental building blocks required for \textit{scalable systems}. This transition from \textit{unstructured collaboration} to \textit{structured componentization} ensures that the reliability of a larger system is mathematically grounded in the verified performance ($S, K \vdash R$) of its atomic units.


\section{Agentic Problem Frames and Agent Development Specification}

This section introduces the \textbf{Agentic Problem Frames (APF)}, a reinterpretation of Michael Jackson’s classical Problem Frames tailored for modern agentic environments, and the \textbf{Agentic Job Description (AJD)}, a foundational specification for agent development derived from the APF. Unlike traditional frameworks for deterministic systems, the APF-AJD framework is designed to systematically accommodate and regulate the inherent stochastic behavior of autonomous agents. By shifting the focus from internal code correctness to the structured interaction between the agent and its environment, this study establishes a systematic engineering methodology to transform probabilistic models into reliable, mission-oriented components.

\subsection{Redefining Entities: Job Performer ($M$) and Workplace ($W$)}

In traditional software engineering, machine ($M$) is conceptualized as a \textit{deterministic system} that executes fixed logic ($S$) based on invariant domain knowledge ($K$) to ensure consistent output. In this paradigm, specifications are static and bound during the design phase. In contrast, an LLM-based agent operates as a \textit{stochastic machine} performing probabilistic inference. Since these models lack fixed deterministic knowledge of specific domains, their unconstrained actions cannot be considered systematic engineering specifications. Instead, the specification takes the form of a \textit{dynamic manifestation} appearing only at runtime.

This study formalizes the generation of the dynamic specification ($S_t$) as follows: 

\begin{equation} (E_t, C_t) \xrightarrow{M} S_t \end{equation}

This formula implies that the execution specification ($S_t$) at time $t$ is the result of the stochastic machine ($M$) processing an incoming event ($E_t$) within the \textit{jurisdictional boundaries} defined by the context ($C_t$). Here, $C_t$ does not merely provide auxiliary information; it functions as a \textit{control framework} that constrains the machine's probabilistic reasoning within a predefined professional domain. By anchoring the machine's behavior through this jurisdictional context, a general-purpose model is transformed into a specialized \textbf{Job Performer} capable of fulfilling specific mission requirements ($R$).

In this framework, the World ($W$) is no longer a static background but functions as an active \textbf{Workplace} where the Job Performer ($M$) executes the mission. While $M$ is the subject of reasoning, the workplace provides the physical and digital infrastructure to \textit{ground} this reasoning. Specifically, the workplace performs two core functions: first, it provides the jurisdictional context necessary to generate $S_t$; second, it produces confirmed knowledge ($K_{t+1}$) by verifying whether the results of actions ($A$) align with ($R$). From an agentic engineering perspective, the workplace is concretized into the following sub-domains:

\begin{itemize}

    \item \textbf{Context and Knowledge Domain ($W_{Context}$):} This is the ``designed domain'' that anchors the agent's probabilistic reasoning. it consists of knowledge databases containing domain-specific facts and manuals, as well as memories of past experiences and success trajectories. Through Retrieval-Augmented Generation (RAG), the workplace injects these materials into $M$, enabling the agent to concretize ambiguous user requests into domain-specific execution specifications ($S_t$). This ensures that the agent's intent is \textbf{bounded} within the organization's jurisdictional rules.
   
    \item \textbf{Interaction Domain ($W_{Interaction}$):} This is the ``causal or biddable domain'' where the agent's capabilities are projected. While traditional PF primarily addressed \textbf{rigidly defined input forms} and \textbf{pre-configured external system connections}, the agentic workplace shifts toward an \textbf{active Tool Use paradigm}. Here, the workplace functions as a \textbf{dynamic gateway} where the agent projects its functional agency into a vast ecosystem. By leveraging standardized protocols such as the \textbf{Model Context Protocol (MCP)}, the agent dynamically invokes external tools to trigger actual state changes. This domain is where abstract specifications ($S_t$) are operationalized into autonomous tool executions ($A$).

    \item \textbf{Verification and Approval Domain ($W_{Verification}$):} This domain acts as an ``epistemological filter'' that addresses the inherent opacity of agentic actions. The completion of an action does not automatically signify mission success. The workplace must verify results through mechanisms such as monitoring callbacks (e.g., confirmation emails) or obtaining explicit user confirmations. Through these, the agent transforms uncertain state changes into \textbf{confirmed knowledge ($K_{t+1}$)}, forming the foundation for the next stage of the mission.
\end{itemize}

\subsection{Governing Agent Execution: Concretized Specification and Verified Knowledge}

The execution of agent $M$ is more than a simple code invocation; it is a governed state evolution process designed to satisfy requirements ($R$) in an uncertain environment. To establish a rigorous engineering boundary for this execution, this study adopts the structure of \textbf{Hoare Logic}, prescribing the pre-conditions and post-conditions that must be met to transform a stochastic action into a reliable engineering event:

\begin{equation} \label{eq:hoare_logic}
    \{S_t\} A \{K_{t+1}\}
\end{equation}

\subsubsection{Pre-condition: Concretization of the Execution Specification ($S_t$)}

In traditional Problem Frames, an action ($A$) relies on complete information and logical rules strictly determined during the design phase. However, in an agentic environment, trigger events—whether user requests or external signals—are inherently ambiguous or incomplete. Therefore, before an agent can enter the execution phase, it must undergo a concretization process that synthesizes these ambiguous inputs with the specific workplace context to manifest a clear \textbf{execution specification ($S_t$)}.

This manifestation process functions not merely as a supplement to missing information but as a core mechanism for \textbf{bounding} the machine's reasoning. By anchoring the stochastic machine ($M$) within the jurisdictional boundaries of a specific job identity, this framework constrains the infinite range of probabilistic outcomes. Through this, the agent evolves from a state of broad capabilities (what it \textit{can} do) to a precise set of professional tasks (what it \textit{must} do) within the given context. Consequently, $S_t$ serves as a rigorous directive that prevents task drift and ensures that the agent's behavior remains consistent with its assigned role.

\subsubsection{Post-condition: Post-Confirmation of Knowledge ($K_{t+1}$)}

Traditional engineering assumes that the results of a machine's execution are reflected in the causal world immediately and deterministically. In an agentic environment, however, the outcomes of an action ($A$) are often opaque due to the machine's stochastic nature and the complexity of environmental variables. Therefore, a change in the state of the world is not considered confirmed simply because an action has been completed. Instead, the post-condition $\{K_{t+1}\}$ must be established through a process of \textbf{post-confirmation} within the verification domain ($W_{Verification}$).

This verification process serves as an essential bridge between the agent's internal reasoning and external reality. By monitoring objective feedback, such as digital callbacks or explicit user approvals, the workplace filters raw action results to determine if the state change aligns with the original intent. Only results that pass this \textbf{confirmation (confirm)} stage are solidified as ``ground truth'' or \textbf{knowledge ($K_{t+1}$)} that the system can safely rely upon.

Ultimately, this ``knowledge-ification'' of results provides the foundational grounding for the next execution cycle. $K_{t+1}$ does not merely signal the end of a task; it provides the verified context necessary to maintain the continuity and reliability of agentic reasoning. By accumulating these confirmed facts, the agent secures an objective basis for subsequent executions,  ensuring reality-based progress toward the requirement ($R$).

\subsection{The AVR Loop: Mechanism for Knowledge Assetization and Convergence}

In agentic systems, the fulfillment of requirements ($R$) is not a one-time event but an evolutionary journey of iterative execution. This study formalizes this cycle as the \textbf{AVR (Act-Verify-Refine) Loop}, representing the core mechanism by which a stochastic machine transforms probabilistic execution into verified domain knowledge assets. This process is governed by the following state evolution logic:

\begin{equation}(E_t, C_t) \xrightarrow{\text{Inject}} S_t \Longrightarrow (S_t, W_t) \xrightarrow{\text{Execute}} W_{t+1} \Longrightarrow W_{t+1} \xrightarrow{\text{Verify}} K_{\Delta} \Longrightarrow (K_t, K_{\Delta}) \xrightarrow{\text{Refine}} K_{t+1} \end{equation}

\subsubsection{Stage 1: Act (Injection and Realization)}

The \textit{Act} stage is a bifurcated process encompassing both the internal manifestation of intent and its external realization.

\begin{itemize}
    \item \textbf{Injection (Concretization):} First, the stochastic machine ($M$) processes an ambiguous trigger event ($E_t$) within the jurisdictional context ($C_t$). This step resolves the inherent incompleteness of the request. By anchoring the event to domain-specific rules and memory, the machine manifests a concrete \textbf{execution specification ($S_t$)}, which is now a rigorous, ``bounded'' set of directives ready for action.

    \item \textbf{Execution (Realization):} Once $S_t$ is manifested, the agent realizes this intent by triggering substantial actions within the workplace ($W_t$). This point marks the transition where internal specifications are applied to the environment, evolving the world into $W_{t+1}$. While $W_{t+1}$ represents a state change, its engineering significance remains opaque until it is formally verified against the original intent.
\end{itemize}

\subsubsection{Stage 2: Verify (Epistemic Determination of $K_{\Delta}$)}

To resolve execution uncertainty, the system invokes a \textbf{\texttt{Verify}} function to observe $W_{t+1}$. This stage serves as an epistemic bridge, extracting \textbf{incremental knowledge ($K_{\Delta}$)}—confirmed facts yielded through action, such as verified API responses or human feedback. This process provides the grounding necessary to confirm whether the state change aligns with the final mission requirements ($R$).

\subsubsection{Stage 3: Refine (From Knowledge Assetization to $K_{t+1}$)}

The \textit{Refine} stage is the heart of knowledge assetization. By integrating the incremental knowledge ($K_{\Delta}$) with the existing knowledge base ($K_t$), the agent establishes an updated state of knowledge, $K_{t+1}$. This represents a structural update to the agent's domain expertise, where $K_{t+1}$ functions as a refined ``map'' of the domain based on the latest execution history.

\paragraph{Propagation to the Next Cycle: From $K_{t+1}$ to $C_{t+1}$}
The updated knowledge ($K_{t+1}$) is not a passive repository; it is immediately propagated as the jurisdictional context ($C_{t+1}$) for the next execution cycle. By feeding refined knowledge back into the machine, the system ensures that the subsequent specification ($S_{t+1}$) is grounded in the verified reality of the previous step. Through this evolutionary progression, the system \textbf{asymptotically converges} toward the mission goal ($R$).

\paragraph{Engineering Significance: Asymptotic Convergence toward $R$}
The significance of the AVR loop lies in its role as a \textbf{closed-loop control system} that drives asymptotic convergence toward mission requirements ($R$). Through the evolutionary cycle of action, verification, and knowledge assetization, the epistemic entropy of internal assumptions is systematically reduced, driving the system from initial probabilistic uncertainty toward deterministic mission requirements. Each iteration transforms a general-purpose probabilistic model into a reliable, professional \textbf{Job Performer}, ensuring that cumulative results align with high-level requirements.

\subsection{Agentic Requirement Satisfaction Argument}

In classical Problem Frames, the satisfaction argument ($S, K \vdash R$) is established through static proof. This assumes that if a perfect specification ($S$) and invariant domain knowledge ($K$) are provided, the requirements ($R$) can be logically and instantaneously derived at the design stage. However, in an LLM-based agentic environment where actions are stochastic and the world is dynamic, this static logic is insufficient.

This study redefines the satisfaction argument as a process of \textbf{asymptotic convergence} driven by the AVR loop. Unlike static proofs, the \textbf{Agentic Requirement Satisfaction Argument} views the fulfillment of requirements as a trajectory of increasing certainty. This argument is formalized as follows:

\begin{equation} \label{eq: limits}
    \lim_{t \to T} \{S_t, K_{t+1}\} \vdash R  
\end{equation}
  
Equation (\ref{eq: limits}) proves that even with a black-box model, reliability is an emergent property of the cumulative verification process rather than an inherent property of the model itself.

In the proposed formulation, $\{S_t, K_{t+1} \}$ represents state variables that evolve over time $t$. The agent incrementally approaches requirement satisfaction through the \textbf{Act-Verify-Refine (AVR) loop}:

\begin{itemize}
    \item \textbf{Concretization of Intent:} To initiate execution, the user's trigger event is constrained by the current domain context ($C_t$) to manifest the execution specification ($S_t$).
    \item \textbf{Epistemic Resolution:} After execution, only results that pass formal verification are solidified and assetized as valid knowledge ($K_{t+1}$).
\end{itemize}

The agent satisfies requirement $R$ through this cumulative sequence of execution and feedback:
\begin{equation}
	(S_0, W_0) \xrightarrow{A_0} K_1, (S_1, W_1) \xrightarrow{A_1} K_2, \dots \to K_{T+1} \vdash R
\end{equation}

This sequence demonstrates that as the total amount of available knowledge ($K$) increases, the \textbf{epistemic uncertainty}—the gap between the agent's assumptions and the actual world state—is systematically resolved. Ultimately, at the termination point $T$, the evidence accumulated in $K_{T+1}$ provides the logical basis for the full satisfaction of goal $R$ through asymptotic convergence.

\subsection{Structural Mechanisms for Agentic Realization}

This section concretizes the detailed mechanisms constituting the agentic satisfaction argument by contrasting them with classical Problem Frames. These mechanisms define the engineering requirements for an agent to function not merely as a software module, but as a \textbf{Job Performer} fulfilling missions within a dynamic workplace.

\subsubsection {Shared Phenomena (Input): Context Injection ($C$)}

In traditional PF, the \textbf{shared phenomena} between the machine ($M$) and the world ($W$) are defined as deterministic, completed events (e.g., pressing an elevator button). In this deterministic world, signals lead to immediate execution without room for interpretation. In contrast, shared phenomena in an agentic environment focus on bridging the semantic gap through an active input process:

\begin{itemize}
    \item \textbf{Existence of a Semantic Gap:} Inputs from users are inherently \textbf{vague intents} (e.g., ``I'm hungry, take me to a restaurant''). A significant semantic gap exists between the abstract intent in the user's mind and the concrete actions the machine must execute.

    \item \textbf{Active Reconstruction via Context Injection:} To bridge this gap, the input interface reconstructs the intent by injecting context. This context consists of the essential information required for the agent to manifest a professional execution specification:

    \begin{itemize}
        \item \textbf{Contexts (Domain Knowledge):} Real-time domain facts, manuals, and jurisdictional regulations infused through RAG. These anchor the agent's reasoning within established professional boundaries. 

        \item \textbf{Memory (Experience):} Re-injection of verified trajectories ($K_{\Delta}$) from previous cycles. This provides an empirical basis for the agent, ensuring it learns from its own operational history and avoids repetitive errors. 

        \item \textbf{Capabilities (Competency):} Technical specifications and constraints of available tools (e.g., MCP). This deterministically limits the scope of the agent's behavioral power to what is actually feasible in the workplace. 
    \end{itemize} 

    \item \textbf{Realization into Execution Specification ($S_t$):} Through the integration of these contextual dimensions, uncertain intent is transformed into a concrete, executable specification. The core of requirement satisfaction lies in this \textbf{input-stage realization}, where the machine accurately interprets the what and how of the mission, manifesting a bounded \textbf{execution specification ($S_t$)} before proceeding to action.
\end{itemize}

\subsubsection{Shared Phenomena (Output): The Epistemic Bridge via Callback and Confirm}

In an agentic environment, a fundamental \textbf{dependency asymmetry} exists between the agent's action ($A$) and the world's reaction. While the machine attempts to change the world, the world does not inherently report the outcomes back into the machine's logical system. This study defines the \texttt{Verify} function as an \textbf{epistemic bridge} to bridge this gap, concretized through a dual-channel of \textbf{Callback (Fact Verification)} and \textbf{Confirm (Value Verification)}.

Our design extends the concept of \textit{Atomic Action Pairs} \citep{thompson2025managing}. While Thompson's \texttt{Guard} functions focus on defensive filters to prevent workflow contamination, our \texttt{Verify} focuses on \textbf{solidifying} any execution result into reliable knowledge ($K_{t+1}$). Even a failure, once confirmed as knowledge, is fed back as a contextual asset ($C_{t+1}$) for the next execution, enhancing system adaptability.

\begin{itemize}
    \item \textbf{Callback (Fact Verification):} A process to synchronize the states of the machine and the world by identifying physical or digital changes. It serves as a lifeline that \textbf{grounds} the agent's reasoning in reality.

    \begin{itemize}
        \item \textbf{Explicit Callback:} Feedback received via paths designed for machine communication (e.g., API responses, Webhooks).
        \item \textbf{Implicit Callback:} In legacy environments without formal paths, the agent detects \textbf{environmental side-effects} (e.g., visual recognition of an elevator's floor display).
    \end{itemize}

    \item \textbf{Confirm (Value Verification):}  This procedure finalizes whether the physical facts identified via callback align with the user's business value and intent. This is achieved through a \textbf{chain of governance}, following the principle that "the executor does not approve their own work." Whether through a human-in-the-loop or a separate critic agent, this final approval ensures \textbf{knowledge finalization}, where only confirmed results are recognized as valid knowledge ($K_{t+1}$) for satisfying requirements ($R$).

\end{itemize}

\subsubsection{Inverse Design and Assetization}

The APF favors an \textbf{inverse design} approach, prioritizing the refinement of external control mechanisms over internal model modification.

\begin{itemize}
    \item \textbf{External Control Primacy:} This study prioritizes \textbf{bounding} and controlling model behavior through the design of external context ($C$) and verification ($V$). Rather than modifying the model's internal parameters, the author focuses on engineering a rigorous environment that steers the stochastic machine toward reliable outcomes.

    \item \textbf{Knowledge Assetization and the Data Flywheel:} Results generated through the agentic loop are transformed into high-quality knowledge assets. The sequences of $\{S_t\} A \{K_{t+1}\}$ constitute \textbf{verified trajectories} that have passed the Confirm stage. These trajectories are fed back into the domain knowledge base ($K$) and jurisdictional context ($C$). This forms a \textbf{data flywheel}—a virtuous cycle where accumulated operational experience refines the ``ground truth'' available to the agent, thereby ensuring higher precision and continuous reliability in future execution cycles.

\end{itemize}


\subsection{Agentic Job Description: Engineering Specification for Job Performers}

The Agentic Job Description is the final engineering output of the APF framework, serving as a systematic specification for agent development. In this study, the AJD is conceptualized as a formal contract between the organization and the agent. This is analogous to a manager assigning a mission to a new employee; to ensure successful performance, the manager must clearly define the mission, the boundaries of authority, the available resources, the professional context to reference, and the explicit criteria for evaluation.

\subsubsection{The AJD as a Blueprint for Specialized Roles}

The engineering significance of the AJD lies in its ability to transform a general-purpose stochastic machine ($M$) into a reliable \textbf{Job Performer} by defining its operational boundaries within the workplace ($W$). The AJD Table (Table \ref{table:ajd_mapping}) functions as a design blueprint that implements the core philosophy of Problem Frames through the following perspectives:

\begin{itemize}

    \item \textbf{Defining Authority and Jurisdiction (Scope \& Workplace)}: Just as a new employee must know their department and reporting line, the AJD defines the \textbf{machine's identity ($M$)} as a professional job performer and its interaction targets within the \textbf{world ($W$)}. It establishes jurisdictional boundaries—explicitly stating what the agent is authorized to do and, more importantly, what it is forbidden to do. This implements PF's focus on defining domain characteristics to ensure the agent’s behavior is bounded within professional limits.

    \item \textbf{Structuring Informational Reference (Operational Context)}: No employee can perform effectively without the right information. The AJD formalizes the "reference materials” required for the job. It defines how vague user intents are concretized into \textbf{execution specifications ($S_t$)} by injecting real-time manuals (context) and previous success trajectories (memory). This ensures that the agent's reasoning is always grounded in the specific workplace reality rather than general probabilistic guessing.

    \item \textbf{Defining Evaluation and Accountability (Evaluation Method)}: In a formal contract, performance is not judged by effort but by results. The AJD prescribes the "epistemic bridge"—the specific callback and confirmation criteria—that must be met to finalize a task. This ensures that the agent's output is verified against objective evidence in the world, satisfying the \textbf{Agentic Requirement Satisfaction Argument} through proven facts.

\end{itemize}

To bridge the gap between abstract engineering and practical organizational design, the AJD components are mapped to the essential elements of a professional job contract for a new employee. Within this framework, \textbf{Mission ($R$)} represents the \textbf{final goal}, explicitly defining the specific outcomes and value expected from the role rather than mere task lists. The \textbf{Workplace ($W$)} identifies the \textbf{office and environment}, specifying the systems, digital tools, and departments the agent is authorized to interact with.

The \textbf{Scope ($M$)} functions as a definition of \textbf{authority and ethics}, establishing the agent’s professional title and, crucially, the "no-go" zones or negative constraints that bound its behavior. For successful execution, the \textbf{Operational Context ($S$)} serves as the \textbf{handbooks and experience}, providing the manuals to be followed and the historical cases to be referenced during reasoning. Finally, the \textbf{Evaluation Method ($K$)} is conceptualized as the \textbf{performance review}, prescribing the objective evidence and verification procedures required to prove the task has been satisfied according to professional standards.

\begin{table}[ht]
\centering
\caption{Mapping Components of the AJD}
\label{table:ajd_mapping}
\small
\renewcommand{\arraystretch}{1.5}
\setlist[itemize]{leftmargin=*, nosep, before=\vspace{-0.2\baselineskip}, after=\vspace{-0.2\baselineskip}}

\begin{tabular}{|>{\RaggedRight\arraybackslash}p{1.6cm}|>{\RaggedRight\arraybackslash}p{1.7cm}|>{\RaggedRight\arraybackslash}p{9.8cm}|} 
    \hline
    \textbf{AJD Component} & \textbf{PF Mapping} & \textbf{Engineering Guidelines \& Implementation Points} \\ \hline
    
    \textbf{Mission} & Requirement ($R$) & 
    The final goal state to be achieved in the world ($W$). \newline
    \textbf{[Result-Oriented]} Focus on "What" rather than "How."
    \begin{itemize}
        \item Define the goal state (e.g., "Prevent accident" instead of "Stop equipment") to allow for reasoning autonomy.
    \end{itemize} \\ \hline
    
    \textbf{Workplace} & World ($W$) & 
    Physical/digital domains the machine interacts with. \newline
    \textbf{[Domain Isolation]} Distinguish if the domain is Causal (API) or Biddable (User).
    \begin{itemize}
        \item Define isolated side-effect boundaries for $W_{User}, W_{Motor}$, etc.
    \end{itemize} \\ \hline
    
    \textbf{Scope} & Machine ($M$) & 
    Identity and jurisdictional boundaries of the machine. \newline
    \textbf{[Machine's Authority]} Specify permissions and \textbf{negative constraints}.
    \begin{itemize}
        \item Use guardrails to prevent hallucinations and job deviation from the professional persona.
    \end{itemize} \\ \hline
    
    \textbf{Operational Context} & Spec ($S$) & 
    Information required to transform events into execution specifications ($S_t$). \newline
    \textbf{[Late Binding]} Inject the following at runtime:
    \begin{itemize}
        \item \textit{Contexts}: Real-time domain knowledge (RAG) and \textbf{Scope/Authority} boundaries.
        \item \textit{Memory}: Re-injection of verified success trajectories ($K_{\Delta}$).
        \item \textit{Capabilities}: Tool specifications (MCP/A2A).
    \end{itemize} \\ \hline
    
    \textbf{Evaluation Method} & Knowledge ($K$) & 
    Epistemic confirmation procedures to prove mission completion. \newline
    \textbf{[Epistemic Grounding]} Design logic to finalize action results as knowledge:
    \begin{itemize}
        \item \textit{Callback}: Fact verification (Physical/Digital sensor data).
        \item \textit{Confirm}: Value verification (Explicit user approval).
    \end{itemize} \\ \hline
\end{tabular}
\end{table}


\section{Case Study: Engineering a Reliable Job Performer}

The purpose of these case studies is to demonstrate how the \textbf{Agentic Job Description} serves as an engineering blueprint to transform human tasks into reliable agentic operations. Rather than simply documenting system requirements, the AJD provides a comprehensive design for \textbf{delegating authority} to a stochastic machine, ensuring that its autonomous judgment consistently aligns with organizational goals.

Two distinct scenarios are presented to illustrate this transformation. First, the \textbf{delegated proxy} model, where the agent acts as a professional assistant navigating complex information systems to resolve ambiguous human intents. Second, the \textbf{autonomous supervisor} model, where the agent functions as a field manager in a physical environment, making independent decisions to maintain system integrity.

Each case illustrates the complete lifecycle of agentic engineering: from modeling the interaction between the performer ($M$) and its workplace ($W$), to defining the boundaries of autonomous judgment that satisfy high-level requirements ($R$). By tracing the agent's journey through these scenarios, this study demonstrates how the AJD framework transforms a general-purpose probabilistic model into a specialized, mission-oriented professional.

\subsection{Case 1: Smart Business Travel Assistant (Delegated Proxy Model)}

The Smart Business Travel Assistant represents a quintessential \textbf{delegated proxy agent} that receives delegated vague intent from a user and performs tasks within the information system domain. The core of this model lies in bridging the semantic gap between the user and the system while minimizing the cognitive cost incurred during complex booking processes.

In this case, the agent's mission extends beyond simple administrative support; it aims to fundamentally eliminate the \textbf{decision fatigue} that employees experience during business travel preparation. By harmonizing corporate regulations with personalized convenience to propose optimal itineraries, the agent functions as a \textbf{cognitive partner}, fostering an environment where employees can remain fully immersed in their core professional responsibilities.

To transform a user’s abstract event, such as ``Prepare for my business trip to Busan next week,'' into a rigorous execution specification ($S_t$), the AJD presented in Table \ref{table:ajd_travel_assistant} is formulated.

\begin{table}[ht]
\centering
\caption{AJD Specification for Smart Trip Assistant}
\label{table:ajd_travel_assistant}
\small
\renewcommand{\arraystretch}{1.5}
\setlist[itemize]{leftmargin=*, nosep, before=\vspace{-0.2\baselineskip}, after=\vspace{-0.2\baselineskip}}

\begin{tabular}{|>{\RaggedRight\arraybackslash}p{1.7cm}|>{\RaggedRight\arraybackslash}p{1.7cm}|>{\RaggedRight\arraybackslash}p{10.3cm}|}
\hline
    \textbf{AJD Item} & \textbf{PF Mapping} & \textbf{Engineering Specification ($S$)} \\ \hline
    
    \textbf{Mission} & Requirement ($R$) & 
    \textbf{[Minimize Administrative Effort]} Minimize direct user intervention and cognitive effort by providing a completed itinerary through personalized scheduling and autonomous reservation. \\ \hline
    
    \textbf{Workplace} & World ($W$) & 
    \begin{itemize}
        \item $W_{User}$: Subject of request and the final value approval domain.
        \item $W_{System}$: Internal email systems, groupware (calendars), etc.
        \item $W_{External}$: Airlines, railway systems, and hotel reservation APIs.
    \end{itemize} \\ \hline
    
    \textbf{Scope} & Machine ($M$) & 
    \begin{itemize}
        \item \textbf{[Identity]}: Senior-level professionalism with deep awareness of corporate travel policies.
        \item \textbf{[Authority]}: Permission for temporary bookings and submission of payment drafts within corporate limits.
        \item \textbf{[Boundary]}: Strict blocking and warning for policy violations (e.g., unauthorized Business class).
    \end{itemize} \\ \hline
    
    \textbf{Operational \newline Context} & Spec ($S$) & 
    \begin{itemize}
        \item \textbf{Contexts}: Meeting schedules and corporate travel reimbursement guidelines.
        \item \textbf{Memory}: Historical preferences (e.g., H-hotel in Haeundae or a preferred window seat).
        \item \textbf{Capabilities}: Real-time API integration for transport and accommodation via MCP.
    \end{itemize} \\ \hline
    
    \textbf{Evaluation \newline Method} & Knowledge ($K$) & 
    \begin{itemize}
        \item \textbf{Callback}: Fact-checking by monitoring email for e-ticket/voucher issuance.
        \item \textbf{Confirm}: Acquisition of explicit final approval from the user on the proposed itinerary.
    \end{itemize} \\ \hline
\end{tabular}
\end{table}

\subsubsection{Domain Modeling via APF Diagram}

\begin{figure}[htbp]
    \centering
    \includegraphics[width=0.8\textwidth]{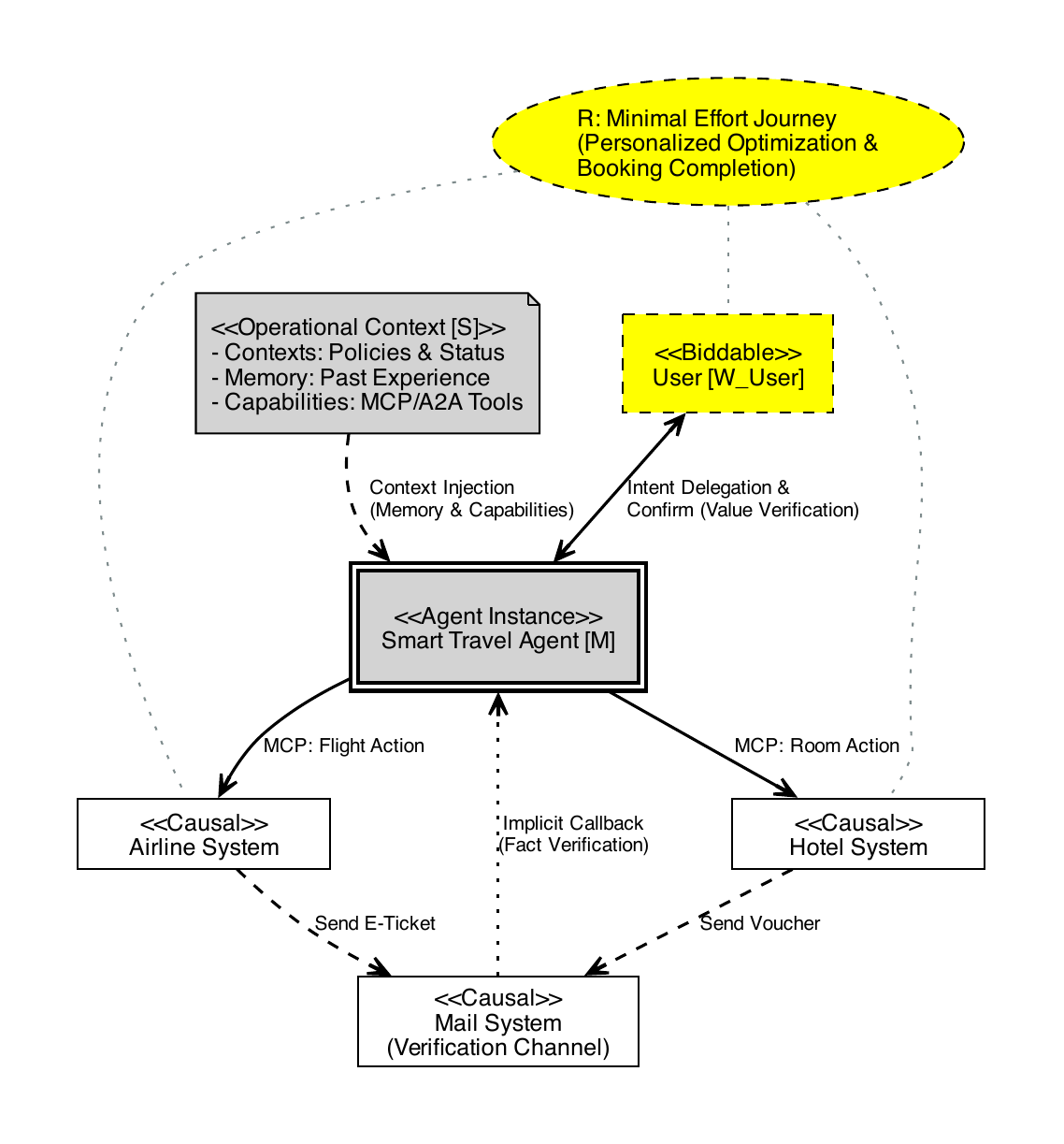}
    \caption{APF diagram for industrial smart travel agent}
    \label{fig:travel_agent}
\end{figure}

Figure \ref{fig:travel_agent} presents the APF diagram for the Smart Business Travel Assistant, serving as a blueprint for the interaction between the performer and its workplace. While maintaining Michael Jackson’s traditional notation, the diagram differentiates visual elements to reflect the agent's characteristics as a stochastic yet bounded entity.

\begin{itemize}

    \item \textbf{Agent Instance ($M$):} The agent $M$, the core of the design, is highlighted with a \textbf{double line}. This signifies that the machine is not merely a computational artifact but an independent performing subject with a job identity and jurisdictional boundaries defined by the AJD.

    \item \textbf{Biddable Domain ($W_{User}$):} The user, represented by a \textbf{dashed line}, is the subject of delegation and value confirmation. This notation specifies that the user's intent is a flexible domain that must be synchronized through dialogue and final approval. 

    \item \textbf{Causal Domains (Airline/Hotel/Mail Systems):} Systems that react deterministically via APIs or protocols (e.g., MCP) are represented by \textbf{solid lines}. These are targets upon which the agent’s capabilities are projected to satisfy the requirement ($R$).
    
    \item \textbf{The Verification Channel (Mail System):} A unique feature of this model is the use of the Mail System as an \textbf{implicit callback} channel. By monitoring the arrival of e-tickets and vouchers, the agent performs fact verification, grounding its logical state in the physical reality of the booking outcome.\footnote{Solid arrows denote direct interactions between the machine and world domains, including requests, intent delegation, and confirmation requests to the user; dotted arrows signify callbacks for notifications or information collection from the world; and dashed arrows represent the flow of information, such as context injection and information transfer between external domains.}

\end{itemize}

\subsubsection{Scenario-based Specification Refinement ($S_t$ Transformation)}

A short utterance from a user, such as ``Prepare for my business trip to Busan next week,'' is a typical incomplete event ($E_t$). At this juncture, the Smart Business Travel Assistant demonstrates its professional expertise by performing the transformation: $(C_t, E_t) \xrightarrow{Inject} S_t$.

The process begins with \textbf{Context Injection}, where the agent, upon receiving the ``Busan'' keyword, refers to corporate travel regulations (e.g., lodging caps, transportation classes) via the Contexts layer. Simultaneously, it checks external environments—such as large-scale conferences in Busan—to recognize inflated hotel prices and explore optimal alternatives within regulations. This is followed by \textbf{Memory Retrieval}, which searches through successful past trajectories to automatically reflect personalized preferences, such as preferred rail seat orientations or specific hotel brands, into the specification. Finally, \textbf{Capabilities Action} is triggered, where the refined specification ($S_t$) invokes MCP-based APIs for airlines, railways, and hotels to perform provisional bookings. This stage moves beyond mere inquiry to realize the essence of delegated booking by organically linking internal payment systems with external APIs.

\subsubsection{Epistemic Verification and Assetization}

In this scenario, the verification procedure is operated concisely, aiming for minimum intervention to reduce the user's cognitive load. The agent undergoes a \textbf{callback} procedure to synchronize its state with physical reality by monitoring the arrival of vouchers and e-tickets from the booking systems. The loop is closed by obtaining the user’s \textbf{confirm} (value approval) for the final proposed itinerary.
Once confirmed, this successful execution trajectory ($K_{\Delta}$) is fed back into the experience  memory. It becomes a core knowledge asset that enables a higher level of automation for future requests, allowing the agent to anticipate needs without requiring explicit instructions. In conclusion, the Smart Business Travel Assistant functions as a proactive agent that minimizes the user’s administrative effort through intelligent context integration, simultaneously satisfying both individual convenience and organizational regulations with minimal intervention.

\subsection{Case 2: Industrial Equipment Manager (Autonomous Supervisor Model)}

The Industrial Equipment Manager represents a quintessential \textbf{autonomous supervisor model}. This agent monitors real-time data from physical sites, independently assesses anomalies, takes immediate corrective actions, and is subsequently evaluated for the appropriateness of its decisions. Unlike the delegated proxy model, this system focuses on fulfilling missions based on the \textbf{Act-then-Report} principle, where timely intervention in physical environments precedes human approval.

To integrate immediate physical control with high-level cognitive judgment, this case employs a \textbf{nested structure}. This architecture combines an edge agent (fast loop) operating at the core of the equipment for low-latency responses, and a central agent (slow loop) that coordinates overall operations and long-term optimization.

\subsubsection{AJD Specification for the Industrial Equipment Manager}

The professional expertise and jurisdictional boundaries of the Equipment Manager are defined in the AJD specification presented in Table \ref{table:ajd_industrial_safety}.

\begin{table}[ht]
\centering
\caption{AJD Specification for Industrial Safety Agent}
\label{table:ajd_industrial_safety}
\small
\renewcommand{\arraystretch}{1.5}
\setlist[itemize]{leftmargin=*, nosep, before=\vspace{-0.2\baselineskip}, after=\vspace{-0.2\baselineskip}}

\begin{tabular}{|>{\RaggedRight\arraybackslash}p{1.6cm}|>{\RaggedRight\arraybackslash}p{1.7cm}|>{\RaggedRight\arraybackslash}p{10.2cm}|}
    \hline
    \textbf{AJD Item} & \textbf{PF Mapping} & \textbf{Engineering Specification ($S$)} \\ \hline
    \textbf{Mission} & Requirement ($R$) & \textbf{[Optimize Safety and Uptime]} Maintain an absolute safe state at all times while minimizing unplanned downtime through proactive intervention and rapid reporting. \\ \hline
    
    \textbf{Workplace} & World ($W$) & 
    \begin{itemize}
        \item $W_{Edge}$: IoT Edge Agent (Intermediary for real-time sensing and control).
        \item $W_{Digital}$: ERP and maintenance management systems (Parts ordering and report storage).
        \item $W_{Manager}$: On-site safety and production managers (Final approvers).
    \end{itemize} \\ \hline
    
    \textbf{Scope} & Machine ($M$) & 
    \begin{itemize}
        \item \textbf{Authority}: Detection of anomalies, emergency intervention (Edge control), root cause analysis, and drafting action reports.
        \item \textbf{Constraint}: Prohibition of unauthorized restarts and final payment for parts without manager approval.
    \end{itemize} \\ \hline
    
    \textbf{Operational \newline Context} & Spec ($S$) & 
    \begin{itemize}
        \item \textbf{Contexts}: Equipment specifications, safety regulations, and failure diagnosis manuals.
        \item \textbf{Memory}: Historical failure patterns and successful response trajectories.
        \item \textbf{Capabilities}: A2A-based Edge communication and ERP system API integration.
    \end{itemize} \\ \hline
    
    \textbf{Evaluation \newline Method} & Knowledge ($K$) & 
    \begin{itemize}
        \item \textbf{Callback}: Physical verification of action results (e.g., confirming RPM=0 via sensor data).
        \item \textbf{Confirm}: Assetization of knowledge through final approval of parts orders and root cause analysis reports.
    \end{itemize} \\ \hline
\end{tabular}
\end{table}

\subsubsection{ Domain Modeling and Scenario Analysis }

\begin{figure}[htbp]
    \centering
    \includegraphics[width=0.8\textwidth]{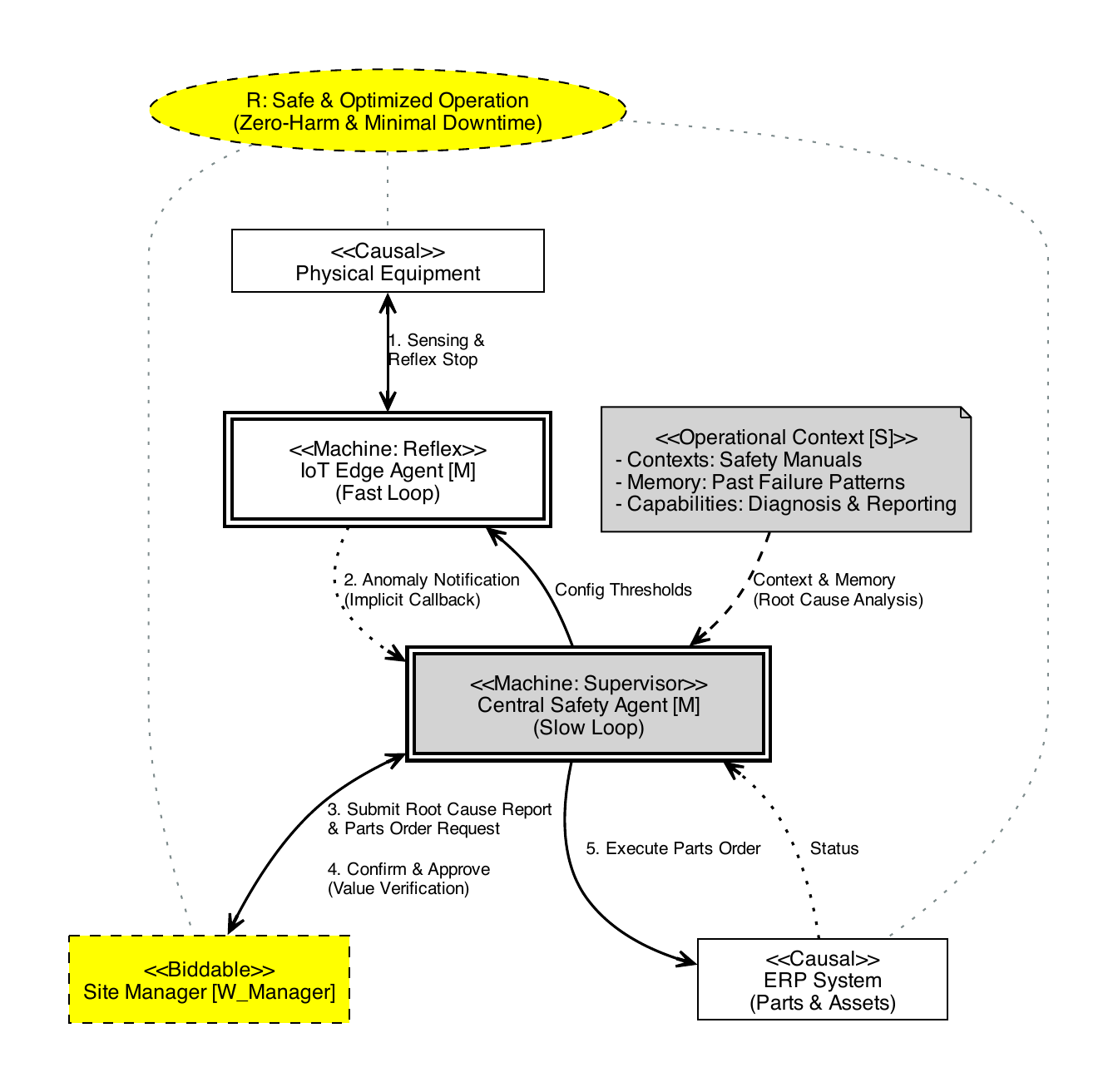}
    \caption{APF diagram for industrial safety manager agent}
    \label{fig:safety_agent_diagram}
\end{figure}

Figure \ref{fig:safety_agent_diagram} presents the APF diagram for the Industrial Equipment Manager. The diagram illustrates a \textbf{nested loop architecture} where the agent instance is decomposed into specialized roles to satisfy requirements in both physical and digital domains.

\begin{itemize}
   \item \textbf{Hierarchical Agent Entities ($M$):} The agent is modeled as a dual-entity system. The IoT Edge Agent (Reflex) handles the "Fast Loop," interacting directly with Physical Equipment (Causal Domain) via sensing and reflex stop. The Central Safety Agent (Supervisor) manages the "Slow Loop," performing high-level reasoning and reporting.
   \item \textbf{The Reporting Chain (Biddable Domain):} The Site Manager, represented as a biddable domain, provides final confirm (Value Verification). This ensures that the agent's autonomous decisions are validated and recorded as organizational knowledge.
   \item \textbf{Causal Interaction (ERP System):} The agent projects its capabilities onto the ERP system to execute parts orders, organically linking safety management with business operations.
\end{itemize}

The Industrial Equipment Manager demonstrates its professional expertise through the following scenario-based transformation:

\begin {itemize}
   \item \textbf{Anomaly Detection and Immediate Intervention:} Upon receiving vibration and temperature anomaly data ($E_t$) from the IoT Edge Agent, the Central Agent diagnoses risks by cross-referencing its injected Context and Memory. If the potential for damage exceeds a predefined threshold, the agent exercises its jurisdictional authority to immediately halt the equipment via $W_{Edge}$, preventing catastrophic accidents.
   \item \textbf{Analysis and Report Generation:} Following intervention, the agent transitions to a reflective stage. It performs root cause analysis, identifies necessary replacement parts, and submits a requisition through the ERP system, preparing follow-up processes to minimize downtime.
   \item \textbf{Value Confirmation and Assetization:} A field manager reviews the report and provides final approval. This Confirm step is a process of epistemic grounding that validates the agent's judgment. Once approved, the response trajectory is fed back into the experience memory, enabling higher precision in future diagnostics.
\end {itemize}

\section{Discussion}

This study extends the traditional Software Engineering (SE) Problem Frames (PF) into agentic environments, proposing a structural foundation to transform stochastic models into reliable engineering components for industrial sites. Beyond the general value alignment of AI, this section discusses the specific principles of control and operation required for an agent to function as a ``trusted colleague'' within a specific domain.

\subsection{From Abstract Quality to Concrete Control through Bounding}

While existing Requirements Engineering for AI (RE4AI) and alignment research have focused on ``universal and abstract qualities'' such as fairness and ethics, the APF proposed in this study addresses quality as an engineering component that is immediately composable and deployable within a specific domain.

In industrial settings, the reliability required of an agent does not depend on the vague level of its intelligence, but on how strictly it fulfills a given \textbf{mission}. To achieve this, APF provides concrete control through \textbf{input/output bounding} as follows:

\begin{itemize}
    \item \textbf{Dynamic Concretization of Input Specification:} Instead of relying on the infinite knowledge of the internet, the agent is constrained to reason within the closed boundaries of domain knowledge ($W_{Context}$) defined by the AJD. Ambiguous natural language requests are transformed into concrete execution specifications ($S_t$) within these boundaries, serving as an engineering fence that prevents the stochastic wandering of intelligence.
    \item \textbf{Assetization and Feedback of Execution Results:} An agent's execution does not end with completion. Incremental knowledge ($\Delta K$) derived through verification ($W_{Verification}$) refines the domain knowledge, which is then fed back as the input context for subsequent executions. Through this process, the agent evolves into a reliable component that \textbf{asymptotically converges} toward the requirements ($R$) over time.
\end{itemize}

Recent findings by \citet{huang2025professional} indicate that developers prefer explicit control through context injection and result verification over the vague creativity of AI. Furthermore, \citet{kurshan2025agentic} emphasized the necessity of self-regulation blocks that can intervene in model execution paths to ensure the safety of financial AI. The APF architecture is a direct result of incorporating these industrial requirements into a formal framework.

\subsection{Clarifying Accountability: From Model Errors to Managerial Responsibility}

When a general-purpose agent produces an incorrect result, the public often attributes the error to the ``AI's mistake,'' shifting accountability into the opaque black box. However, in an APF-based domain agent environment, the locus of accountability must be redefined as a matter of the \textbf{adequacy of context injection and specification}.

Viewing the agent as a professional \textbf{Job Performer}, failures can be explained through the following distinct engineering perspectives:
\begin{itemize}
    \item \textbf{Insufficient Jurisdictional Context:} The failure to provide the necessary information required for execution.
    \item \textbf{Knowledge Mismatch:} The injected domain knowledge was inconsistent with reality, leading to the generation of flawed input specifications ($S_t$).
    \item \textbf{Deficient Criteria:} The failure to define objective success conditions required to verify and evaluate mission fulfillment.
\end{itemize}

Ultimately, the issue of agent accountability shifts from a psychological inquiry (e.g., ``Why did the AI misjudge?'') to a \textbf{managerial and engineering responsibility} (e.g., ``Did the system provide sufficient authority, context, and clear evaluation criteria for the agent to complete the mission?''). This approach moves away from the irresponsibility of blaming the black box and provides a practical metric for accountability based on compliance with specification and verification procedures.

\subsection{Redefining Operation: Adaptation through Reality Sync}

This study redefines the operation of agentic systems as a continuous process of managing the gap between the machine’s ($M$) internal ``assumptions about the world'' and the ever-changing ``reality.'' Since the underlying foundation models provide only universal reasoning capabilities, real-time field information must be continuously injected into the agent's context to satisfy specific industrial requirements over time.

In this process, the post-verification procedure functions beyond a simple pass/fail assessment; it acts as a \textbf{critical sensor} detecting changes in the physical or digital world. Detected environmental shifts immediately update the knowledge base linked to the AJD, allowing the agent to flexibly adapt to changing realities through updated context without requiring a separate retraining process.

Consequently, the true essence of agentic engineering lies not in increasing the ``size'' of intelligence, but in how robustly one structures the grounding upon which the agent stands, and how up-to-date that grounding is maintained through the loop of execution and verification. The APF and AJD proposed in this study serve as the minimum specifications and protocols that enable such engineering control.

This perspective aligns with contemporary research. \citet{hu2025memory} emphasized the growth of context ($C$) through the evolution of \textbf{experience memory}, while \citet{schick2023toolformer} demonstrated tool-use optimization via \textbf{Toolformer}, and \citet{shinn2023reflexion} proved the possibility of machine learning through self-reflection in \textbf{Reflexion}. While these efforts enhance agent adaptability at different layers, the APF provides the \textbf{architectural foundation} where these individual technologies can orderly converge toward the engineering goal of ``Requirement Satisfaction.''

\section{Conclusion}

This research proposes the Agentic Problem Frames as an engineering foundation for transforming general intelligence into a trusted ``colleague'' within specific domains. Moving beyond the conventional view of agents as opaque, intelligent black boxes, this study redefines them as independent, composable engineering components that can be integrated into complex organizational systems.

The most pivotal differentiation established in this research lies in the paradigm of specification. Unlike traditional software that relies on fixed logic, specifications for stochastic machines like agents cannot be finalized at the design stage. Instead, the author presents a \textbf{dynamic specification ($S_t$)} paradigm, where ambiguous natural language requests are synthesized with domain knowledge ($K$) to manifest as concrete directives just before execution.

To guide the agent’s probabilistic reasoning toward deterministic requirement satisfaction ($R$), the proposed framework internalizes the \textbf{AVR (Act-Verify-Refine) feedback loop} as an architectural device. This loop enables the \textbf{assetization} of execution results by verifying outcomes and re-injecting them as context for subsequent iterations. Through this structural control, reliable unit agents—governed by clear Agentic Job Description—can be composed into higher-order systems capable of executing the large-scale, complex missions of an organization.

In conclusion, the reliability of agents in industrial settings stems not from the model's intelligence (IQ), but from the engineering structuring that refines knowledge through repeated execution and verification. This study formalized this essential convergence process with the following equation:

$$\lim_{t \to T} \{S_t, K_{t+1}\} \vdash R$$  

This formula proves that when the specification ($S_t$) at each time step is continuously calibrated by verified knowledge assets ($K_{t+1}$), the agent inevitably reaches the point of trust ($R$). The author expects the framework presented in this study to serve as a robust milestone, grounding agents as legitimate members of the industrial workforce rather than mere technological novelties.

\section*{Call for Collaboration: Empirical Validation}

The author is actively seeking partners to apply the proposed APF and AJD frameworks to complex, real-world industrial domains to further validate their efficacy. I welcome researchers and practitioners who wish to implement agents as composable components to achieve organizational missions or to empirically validate the process of incremental requirement satisfaction within their own domains.

\textbf{Contact:} [calling6@snu.ac.kr]

\bibliographystyle{unsrtnat}
\bibliography{references}

\end{document}